\documentclass{article}
\usepackage{PRIMEarxiv}
\usepackage{natbib}
\usepackage[utf8]{inputenc}
\usepackage[T1]{fontenc}    % use 8-bit T1 fonts
\usepackage{xr-hyper}
\usepackage{hyperref}       % hyperlinks
\usepackage{url}            % simple URL typesetting
\usepackage{booktabs}       % professional-quality tables
\usepackage{amsmath, amsfonts}       % blackboard math symbols
\usepackage{nicefrac}       % compact symbols for 1/2, etc.
\usepackage{microtype}      % microtypography
\usepackage[dvipsnames, table]{xcolor}
\usepackage{caption}
\usepackage{subcaption}
\usepackage{graphicx}
\usepackage{enumitem}
\usepackage[ruled,vlined]{algorithm2e}

\graphicspath{{Images/}}
\DeclareGraphicsExtensions{.pdf,.png} 
\DeclareMathAlphabet\mathbfcal{OMS}{cmsy}{b}{n}

\makeatletter
\newcommand*{\addFileDependency}[1]{% argument=file name and extension
  \typeout{(#1)}
  \@addtofilelist{#1}
  \IfFileExists{#1}{}{\typeout{No file #1.}}
}
\makeatother

\title{Train on Small, Play the Large: Scaling Up Board Games with AlphaZero and GNN}
\author{
    Shai Ben-Assayag \\
    \texttt{sbassayag@cs.technion.ac.il} \\
  \And
    Ran El-Yaniv \\
    \texttt{rani@cs.technion.ac.il}
}
\date{May 2021}

\begin{document}

\maketitle
\begin{abstract}
Playing board games is considered a major challenge for both humans and AI researchers. Because some complicated board games are quite hard to learn, humans usually begin with playing on smaller boards and incrementally advance to master larger board strategies. Most neural network frameworks that are currently tasked with playing board games neither perform such incremental learning nor possess capabilities to automatically scale up. In this work, we look at the board as a graph and combine a graph neural network architecture inside the AlphaZero framework, along with some other innovative improvements. Our ScalableAlphaZero is capable of learning to play incrementally on small boards, and advancing to play on large ones. Our model can be trained quickly to play different challenging board games on multiple board sizes, without using any domain knowledge. We demonstrate the effectiveness of ScalableAlphaZero and show, for example, that by training it for only three days on small Othello boards, it can defeat the AlphaZero model on a large board, which was trained to play the large board for $30$ days.

\end{abstract}

\section{Introduction}
\label{introduction}
Learning a simple instance of a problem with the goal of solving a more complicated one is a common approach within various fields. Both humans and AI programs use such incremental learning, particularly when the large-scale problem instance is too hard to learn from scratch or too expensive. This paper is concerned with applying incremental learning to the challenge of mastering board games. When playing board games, humans have the advantage of being able to learn the game on a small board, recognize the main patterns, and then implement the strategies they have acquired, possibly with some adjustments, on a larger board. In contrast, machine learning algorithms usually cannot generalize well between board sizes. While simple heuristics, such as zero padding of the board or analyzing local neighborhoods, can alleviate this generalization problem, they do not scale well for enlarged boards (see, e.g., Section~\ref{Evaluation}). 

%The challenge of mastering board games can benefit from incremental learning as well. Learning to play board games that have large state space and branching factors often requires extremely complex tactics. Despite being solved completely for small-scale boards (e.g., by exhaustive search), when dealing with computers, this is still considered highly challenging for large ones. 

%In addition, zero padding still requires that both training and playing are performed with respect to a predefined board size, and thus do not enjoy the potential computational benefits of scalable methods. 

In this paper we propose ScalableAlphaZero (SAZ), a deep \emph{reinforcement learning} (RL) based model that can generalize to multiple board sizes of a specific game. SAZ is trained on small boards and is expected to scale successfully to larger ones.
Our technique should be usable for \textit{scalable board games}, whose rules for one board size apply to all feasible board sizes (typically, infinitely many). 
%which are well-definedi.e., games that can be played on infinitely many assorted sized boards and do not necessitate any update to the formal rules of the game
For instance, Go is scalable but standard chess is not. A strong motivation for finding such a model is a potential substantial reduction in training time. As we demonstrate in this paper, training a model on small boards takes an order of magnitude less time than on large ones.
The reason is that the dimension of states is significantly smaller, and gameplay requires fewer turns to complete. 
%Moreover, one scalable model 
%another motivation is the versatility of the scalable player. 
%our analysis show that by training only one adaptable model the player 
%can play consistently and successfully on multiple board sizes.

%OLD: The motivation for finding a scalable RL model is driven by two objectives: a) \textbf{reducing training time} and b) \textbf{gaining flexibility}. Clearly, training our model on small board sizes takes much less time than on large ones, as gameplay requires fewer turns to complete and the space of training examples is significantly smaller. This small-scaled learning approach also enables the system to operate entirely using modest computational resources. Flexibility is gained by training only one model, which later on can be used to play any other board size. 

The proposed model is based on two modifications of the well-known AlphaZero (AZ) algorithm \citep{silver2017mastering_2}. To the best of our knowledge, presently AZ is the strongest superhuman RL based system for two-player zero-sum games. The main drawback of AZ is that it limits the user to training and playing only on a specific board size. This is the result of using a convolutional neural network (CNN) \citep{atlas1987artificial} for predictive pruning of the AZ tree. To overcome this obstacle, in SAZ we replace the CNN by a graph neural network (GNN) \citep{scarselli2008graph}. The GNN is a \textit{scalable} neural network, i.e., it is an architecture that is not tied to a fixed input dimension. GNN's scalability enables us to train and play on different board sizes and allows us to scale up to arbitrarily large boards with a constant number of parameters. To further improve the AZ tree search pruning, we propose an ensemble-like node prediction using \emph{subgraph sampling}; namely, we utilize \textbf{the same} GNN for evaluating a few subgraphs of the full board and then combine their scores to reduce the overall prediction uncertainty.

We conduct experiments on three scalable board games and measure the quality of SAZ by comparing it to various opponents on different board sizes. Our results indicate that SAZ, trained on a maximal board size of $9\times 9$, can generalize well to larger boards (e.g., $20\times 20$). Furthermore, we evaluate it by competing against the original AZ player, trained on a large board. Our model, with around ten times less training (computation) time on the same hardware, and without training at all on the actual board size that was used for playing, performs surprisingly well and achieves comparable results.

The main contributions of this work are: (1) a model that is capable of successfully scaling up board game strategies. As far as we know this is the first work that combines RL with GNNs for this task; (2) a subgraph sampling technique that effectively decreases prediction uncertainty of GNNs in our context and is of potential independent interest; (3) the presentation of extensive experiments, demonstrated on three different board games, showing that our model requires an order of magnitude less training time than the original AZ but, still, can defeat AZ on large boards.

\section{Related work} 
\label{sec:related}
The solution proposed in this paper instantiates a GNN model inside the AlphaZero model for the task of scalable board game playing. In this section, we briefly review early work in AI and board games, focusing on the AlphaZero \citep{silver2017mastering_2} algorithm. We further describe the GNN design and review various works that use GNN to guide an RL model. Finally, we summarize existing methods that aim to deal with scalable board games and accelerate the generalization between sizes.

\subsection{AlphaZero for board games}
\label{related:RL}
Given an optimization problem, deep RL aims at learning a strategy for maximizing the problem's objective function. The majority of RL programs do not use any expert knowledge about the environment, and learn the optimal strategy by exploring the state and action spaces with the goal of maximizing their cumulative reward.

AlphaGo (AG) \citep{silver2016mastering} is an RL framework that employs a policy network trained with examples taken from human games, a value network trained by selfplay, and Monte Carlo tree search (MCTS) \citep{coulom2006efficient}, which defeated a professional Go player in 2016.
About a year later, AlphaGo Zero (AGZ) \citep{silver2017mastering} was released, improving AlphaGo's performance with no handcrafted game specific heuristics; however, it was still tested only on the game of Go. AlphaZero \citep{silver2017mastering_2} validated the general framework of AGZ by adapting the same mechanism to the games of Chess and Shogi. AG and AGZ have a three-stage training pipeline: selfplay, optimization and evaluation, whereas AZ skips the evaluation step. AGZ and AZ do not use their neural network to make move decisions directly. Instead, they use it to identify the most promising actions for the search to explore, as well as to estimate the values of nonterminal states.

\subsection{Graph neural networks}
\label{related:GNN}
GNNs, introduced in \cite{scarselli2008graph}, are a promising family of neural networks for graph structured data. GNNs have shown encouraging results in various fields including natural language processing, computer vision, logical reasoning and combinatorial optimization. Over the last few years, several variants of GNNs have been developed (e.g., \cite{hamilton2017inductive, gilmer2017neural, li2015gated, velivckovic2017graph, defferrard2016convolutional}), while the selection of the actual variant that suits the specific problem depends on the particularities of the task.

In their basic form, GNNs update the features associated with some elements of an input graph denoted by $G=(V,E)$, based on the connections between these elements in the graph. A \textit{message passing} algorithm iteratively propagates information between nodes, updates their state accordingly, and uses the final state of a node, also called ``node embedding'', to compute the desired output. Appendix~\ref{supp:MP} provides more details about the message passing procedure. In this paper we use graph isomorphism networks (GINs) \citep{xu2018powerful}, which are a powerful well-known variant of GNNs. For further details about GINs, see Appendix~\ref{supp:GIN}. 

\subsection{Scalable deep reinforcement learning}
\label{scalableRL}
Recently, several works tackled the problem of scalability in RL in the context of combinatorial optimization using GNNs that are natural models to deal with such challenges. For example, \cite{lederman2018learning} utilized the REINFORCE algorithm \citep{williams1992simple} for clause selection in a QBF solver using a GNN, and successfully solved arbitrary large formulas. \cite{abe2019solving} combined Graph Isomorphism Networks \citep{xu2018powerful} and the AGZ framework for solving small instances of NP-complete combinatorial problems on graphs. \cite{dai2017learning} proposed a framework that combines RL with structure2vec graph embedding \citep{dai2016discriminative}, to construct incremental solutions for Traveling Salesman and other problems. Other RL models that deal with combinatorial optimization problems include \cite{yolcu2019learning,xing2020graph}.

A fundamental difference between trying to scale combinatorial optimization problems and our task is that a reductionist approach is much less intuitive for scaling up board games. For example, when trying to solve a large-scale SAT instance (as in \cite{yolcu2019learning}), the problem necessarily gets smaller as long as the search advances. More specifically, by setting a literal $x_i$ to $True$ or $False$, all clauses that contain $x_i$ or $\neg{x_i}$ can be deleted (either in conjunctive normal form or disjunctive normal form). In contrast, in a board game, the problem size remains the same during the entire search, with much more challenging rare boards that have not yet been encountered.

Among existing work on using learning to scale up board games, the most similar to our approach is that of \cite{schaul2009scalable}. To enable size generalization for Go-inspired board games, they presented MDLSTM, a scalable neural network based on MDRNNs and LSTM cells, computing four shared-weight swiping layers, one for each diagonal direction on the board. For each position on the board, they combine these four values into a single output representing the action probabilities. Their results show that MDLSTM transfers the strategies learned on small boards to large ones, leading to a level of play on a $15\times15$ board that is on par with human beginners. Some other similar approaches include those of \cite{gauci2010indirect, wu2007scalable}. \citeauthor{gauci2010indirect} extrapolated $5\times 5$ Go solutions to $7\times7$, thus speeding up the training. \citeauthor{wu2007scalable} designed a DAG-RNN for Go and demonstrated that systems trained using a set of $9\times9$ amateur games achieve surprisingly high correlation to the strategies obtained by a $19\times19$ professional players' test set.
 
All the above models, aimed at scaling up board games, do not incorporate an RL framework within their model, neither for training nor playing. In contrast to our model, which starts its training as a \textit{tabula rasa} (i.e., without using any specific domain knowledge), the training processes of \citeauthor{schaul2009scalable} and \citeauthor{gauci2010indirect} are based on playing against a fixed heuristic based opponent, while \citeauthor{wu2007scalable} trained their model using records of games played by humans.
 
\section{Scalable AlphaZero for board games}
\label{description}
In this section we describe in detail our RL based model for scalable board games. Our model is based on AZ, equipped with additional components that allow it to train on small board sizes and play on larger ones. The board game environment encodes the rules of the game and maintains the board state. We denote by $\mathcal{A}$ the set of possible actions and by $\mathcal{S}$ the set of possible board states.

As mentioned in Section~\ref{related:RL}, the AZ player is an RL model consisting of a combined neural network,
$ f_\theta:\mathcal{S} \rightarrow [0,1]^{|\mathcal{A}|}\times \{-1,0,1\} $, with parameters $\theta$ and an MCTS. The network takes as input the raw board representation of the current state $s\in \mathcal{S}$, and outputs $f_\theta(s)=(\mathbf{p}_s,v_s)$, where the probability vector  $\mathbf{p}_s=(p_1,p_2,...,p_{|A|})\in [0,1]^{|\mathcal{A}|}$ represents the probabilities of selecting each action on the board, and the value $v_s\in [-1,1]$ estimates the chances of the current player winning the game (i.e., $-1$ for losing, $0$ for a tie and $1$ for winning), given its current state. At each state $s$, an $f_\theta$-guided MCTS is activated. 
The MCTS procedure then outputs the probability $\pi$ for playing each valid move. For a full description of the MCTS procedure, see Appendix~\ref{supp:MCTS}.
The pseudocode for our model, including the MCTS procedure, is provided in Appendix~\ref{supp:Pseudocode}. 

To summarize, the main changes we made to the original AZ are%: (1) replacing the CNN by our GNN; (2) adding subgraph sampling for guidance of the MCTS procedure; and (3) removing rotation and reflection augmentations in the training set.
\begin{itemize}[leftmargin=1cm]
    \item Replacing the CNN by our GNN.
    \item Adding subgraph sampling for guidance of the MCTS search.
    \item Removing rotation and reflection augmentations in the training set.
\end{itemize}
The next sections elaborate on each of these components.

\subsection{Replacing the CNN}
The main difference between our scalable RL player and AZ comes from choosing the specific neural network type. AZ uses a CNN as the network $f_\theta$. As already mentioned, CNN architectures are limited due to the specific input they require, thus they do not enjoy the potential computational benefits of scalable methods. The message passing technique used in a GNN \citep{gilmer2017neural} (see Section~\ref{related:GNN}) allows the network to get a variable sized graph with no limitation on either the number of nodes or the number of edges. In fact, a GNN only requires a fixed size of feature dimension for each node (and each edge, if edge features are used). This last observation makes a GNN a scalable neural network according to the definition above. Consequently, replacing the original CNN in the AZ framework with a GNN is a key step toward our construction of a scalable player mechanism.

\begin{figure}[htp]
    \centering
    \begin{subfigure}[b]{0.32\textwidth}
        \centering
        \includegraphics[width = 0.6\textwidth]{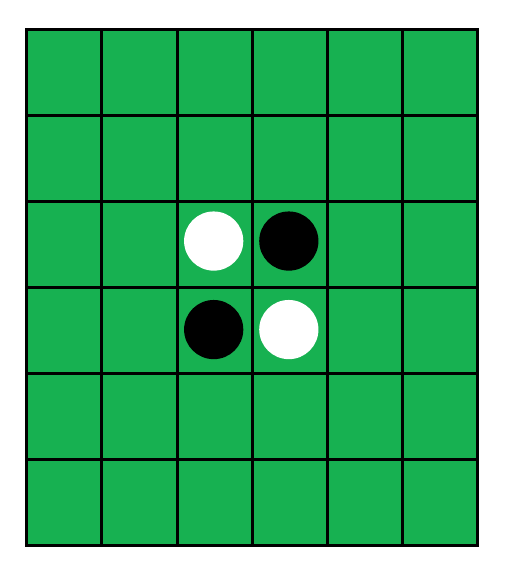}
        \caption{Othello initial board of size $6\times 6$}
        \label{fig:othelloInit}
    \end{subfigure}
    \begin{subfigure}[b]{0.38\textwidth}
        \centering
        \includegraphics[width = \textwidth]{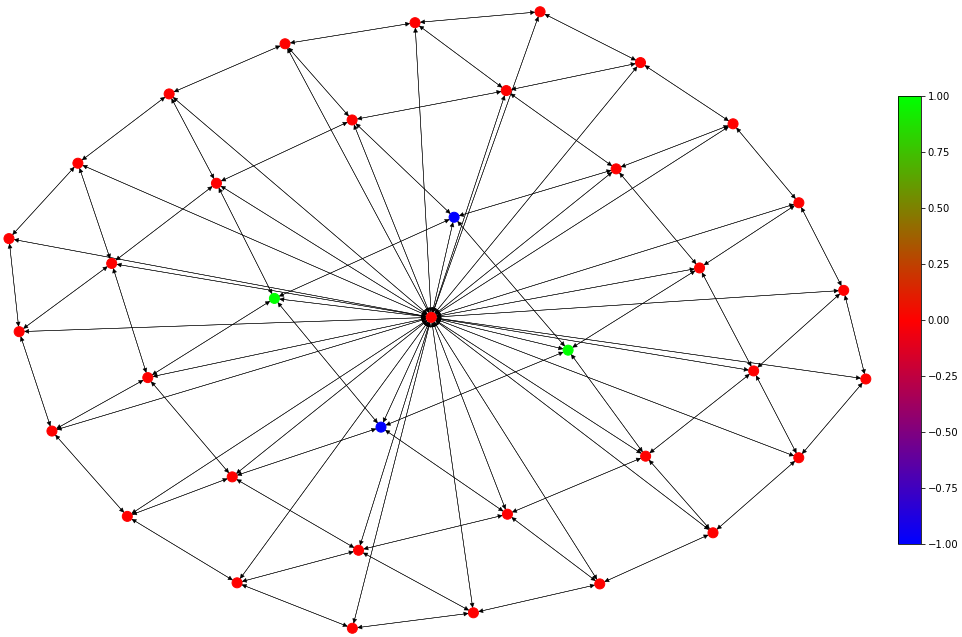}
        \caption{Corresponding graph}
        \label{fig:graph}
    \end{subfigure}
    \caption{Illustration of the graph $G(s_0)$ generated based on the initial board state, $s_0$, of the game Othello. The blue and green nodes correspond to the light and dark pieces, respectively. Our additional dummy node is the central red with black border node connected to all other nodes. It is the only node that does not represent any square on the board.}
    \label{fig:builtGraph}
\end{figure}

To instantiate $f_\theta$ as a GNN, we first need to translate the board state into a graph. We define the graph $G(s)=(V,E)$ where nodes in $V$ are the positions on the board (usually, $V=\{(i,j) | i,j\in [n]\}$ for a grid-like square board of size $n\times n$),
and the edges in $E$ connect ``geographically'' adjacent positions on the board (for the grid-like example above we connect only vertical and horizontal neighbors and discard diagonal neighbors). For a node $v\in V$ we denote by $h^0_v$ the initial feature representing the current piece placed on $v$ ($-1$ for a light piece, $1$ for a dark piece and $0$ for an empty square). 
Last, we add a \textit{dummy node} (as demonstrated in \cite{gilmer2017neural}) that is connected to all other nodes in $V$, allowing us to improve the long-distance data flow between nodes. The dummy node has an initial feature $h^0_\textrm{dummy}=0$. 
Figure~\ref{fig:builtGraph} illustrates the graph generation procedure, which corresponds to the initial board of the Othello game.

\begin{figure}[htp]
    \centering
    \includegraphics[width=0.9\linewidth]{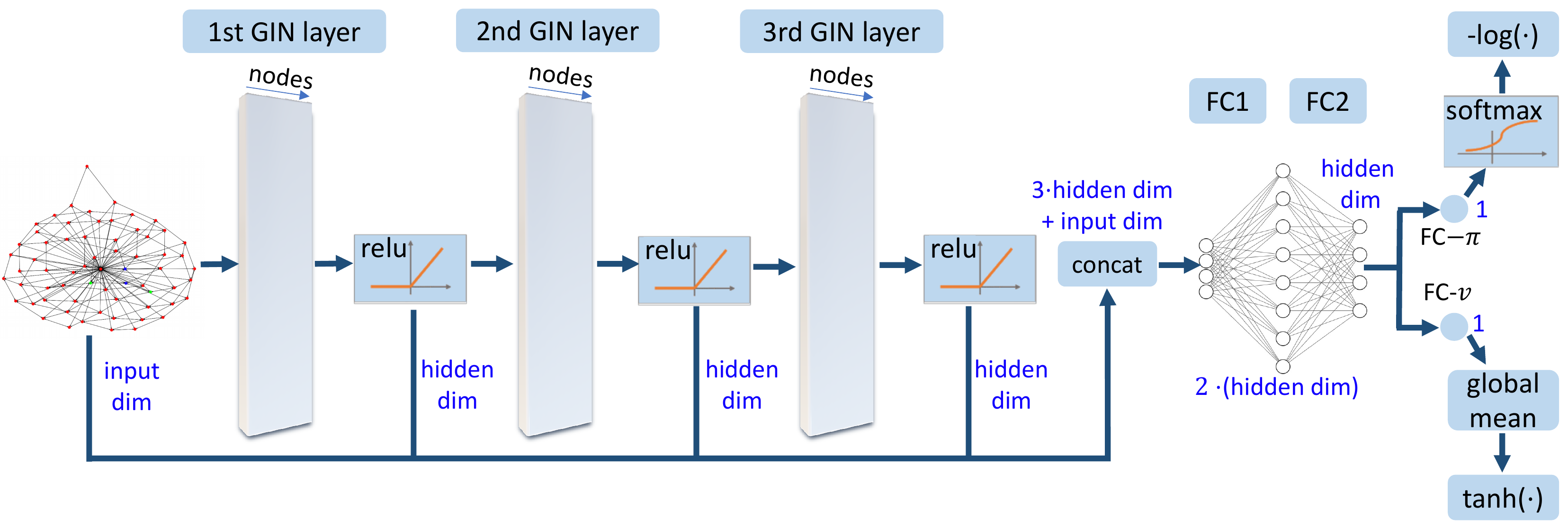}
    \caption{Neural network architecture}
    \label{fig:architecture}
\end{figure}

Our GNN receives the generated graph as input and outputs both the probability for playing the specific action corresponding to the node and the value of the current state (i.e., the whole graph). The final GNN architecture, which is based on the GIN model (see the discussion in Section~\ref{related:GNN}) with extra skip connections, is illustrated in Figure~\ref{fig:architecture}. 
The architecture was implemented using PyTorch Geometric \citep{fey2019fast}. It contains the following modules:
\begin{enumerate}[leftmargin=*]
    \item Three GIN layers with layer normalization and a $ReLU$ activation function.
    \item Concatenation of all previous intermediate representations.
    \item Two fully-connected layers with batch normalization, $ReLU$ activation function and dropout.
    \item The computation is separated into two different heads, for computing the policy $\mathbf{p}$ and the value $v$. $\mathbf{p}$ is computed using one fully-connected layer, followed by a $\log$-$softmax$ operation, yielding the probability vector. $v$ is computed using one fully-connected layer, followed by a global mean pooling layer (i.e., the mean among all nodes) and, finally, a $\tanh$ nonlinearity function.
\end{enumerate}

\subsection{Guiding MCTS}
\label{guidingMCTS}
The second change we made refers to the guidance of the MCTS by the network $f_\theta$. According to MCTS, $f_\theta(s')=(\mathbf{p_{s'}},v_{s'})$ is computed for each nonterminal leaf node $s'$ discovered during the game. These values are used for updating the MCTS variables $P(s',\cdot)=\mathbf{p}_{s'}$, propagating $v_{s'}$ along the path seen in the current game simulation, and updating their $Q(s,a)$ values accordingly.

Here we can take advantage of the scalability of our network $f_\theta$, and enhance the performance of the tree search. Upon arriving at $s'$, we sample a few subgraphs of the graph generated by $s'$ and send them to $f_\theta(\cdot)$. For each subgraph we first sample the subgraph size $d\in[(m-1)^2,m^2]$ and then sample $d$ nodes present in the subgraph. The subgraph size $d$ should be large enough to form an ``interesting'' new state and include enough legal actions. The subgraphs' size range ($m$) as well as the number of sampled subgraphs are two hyperparameters of our model. Note that sending more than one graph to the network for each newly visited leaf node can be implemented efficiently using batches, which increases the prediction time by only a small factor. Our experiments show that using a small number of subgraphs improves the player's performance remarkably.

The MCTS variables are updated in our model according to
$
P(s',\cdot)=\nicefrac{ \left ( \mathbf{p_1} + \mathbf{p_1} \circ \mathbf{p_2} \right ) }{2}
$,
where $\mathbf{p_1}$ is the probability vector $\mathbf{p}_{s'}$ taken from the evaluation $f_\theta(G(s'))$, $\mathbf{p_2}$ is the \emph{scatter mean/max} of the probability vectors computed on the subgraphs (i.e., it takes into account how many times a node was sampled), and $\circ$ stands for element-wise multiplication. Propagating $v(s')$ remains unchanged.\footnote{GitHub repository: \href{https://github.com/rusty1s/pytorch_scatter}{pytorch\_scatter} (released under the MIT license)} 

\subsection{Training pipeline}
\label{trainingPipeline}
The training pipeline, as in the AZ model, comprises a loop between the \textit{selfplay} and optimization stages. The game result, $z\in \{-1,0,1\}$, of each selfplay is propagated to all the states visited during the game. The player plays against itself, thus accumulating positive and negative examples. The neural network parameters are optimized at the end of the selfplay stage to match the MCTS probabilities $\pi$ and the winner $z$. For more details about the AZ training pipeline, see Appendix~\ref{supp:pipeline}.

For each training example produced during selfplay, AGZ generates extra examples by looking at rotations and reflections of the board. In contrast, AZ did not use these extra training examples, demonstrating the strength of their guiding network. By looking at the board as a graph, our GNN takes these invariances into account, thus justifying the removal of extra examples without the need to enhance the performance of the guiding network (e.g., by increasing the number of parameters). Consequently, removing rotation and reflection examples results in a massive reduction in the required training resources and substantially speeds up training time (by 5x).

\section{Evaluation}
\label{Evaluation}
We conduct our experiments on three scalable board games: (1) \textbf{Othello} \citep{landau1985othello}: also known as Reversi. Players alternately place stones on the board trying to ``capture'' the opponent's stones. Any straight line sequence of stones belonging to the opponent, lying between the just placed stone and another stone of the current player, are turned over and switch colors. The winner is determined by the majority stones' color.
%\item \textbf{Atari-Go} : Also known as Ponnuki-Go or ‘Capture Game’, is a simplified version of Go \citep{smith1908game} that is widely used for teaching the game of Go to new players. The rules are the same as for Go, except that passing is not allowed, and the first player to capture a predetermined number (here: one) of his opponent’s stones wins.
(2) \textbf{Gomoku}: also known as ‘Five in a row’ or Gobang. Players take turns placing stones on the board. The first player to place $k$ (here $5$) stones in a row, a column or a diagonal, wins.
(3) \textbf{Go:} the well-known game of Go \citep{smith1908game}. Two players alternately place stones on intersections of the board with the goal of surrounding more territory than the opponent.
Table~\ref{table:gameComplexity} analyzes the game complexity of the games used for testing.

\begin{table}[ht!]
\caption{Strategic complexity of small/large Othello, Gomoku and Go games given by evaluations of their state and action space size (upper bound).} 
\label{table:gameComplexity}
\centering
\begin{tabular}{ lrrrr  }
  \\
  \toprule
  & \multicolumn{2}{c}{\textbf{Othello}} & \multicolumn{2}{c}{\textbf{Gomoku/ Go}} \\
  \cmidrule(r){2-3} \cmidrule(r){4-5}
  & \multicolumn{1}{c}{$\mathbf{8\times8}$} & \multicolumn{1}{c}{$\mathbf{16\times16}$} & \multicolumn{1}{c}{$\mathbf{9\times9}$} & \multicolumn{1}{c}{$\mathbf{17\times17}$} \\
  \midrule
  $|\mathbfcal{S}|$ (states) & $262,144$ & $16,777,216$ & $531,441$ & $24,137,569$ \\
  $|\mathbfcal{A}|$ (actions) & $65$ & $257$ & $82$ & $290$ \\
  \bottomrule
\end{tabular}
\end{table}

We define two \textit{reference opponents} for each game: a \textit{random player} that randomly chooses a legal move, and a \textit{greedy player} that chooses his action based on a hand-coded tactical heuristic score. The specific heuristics for each game is described in Appendix~\ref{supp:heuristics}. The greedy opponent provides a sufficient challenge to demonstrate the utility of generalization. Note that both reference players can play on every board size without making any changes to the action-choosing mechanism.

As a measure of success we use the \textit{average outcome}\label{outcome} of 100 games against one of the reference opponents, counted as $1$ for a win, $0.5$ for a tie and $0$ for a loss. Each player plays half the time with dark pieces (plays first) and half with light pieces (plays second).
We also analyze individually each main change we made. Furthermore, we play against the original AZ player that was trained to play on a large board, which enables us to measure the effect of our improvements on the training speed and realtime playing performance. Full CNN architecture of the AZ player in described in Appendix~\ref{supp:cnnArch}. All tables and graphs provided include standard errors (five independent runs).

\subsection{Experimental setup}
\label{setup}
Our RL infrastructure runs over a physical computing cluster. To train SAZ, we use one GPU (TITAN X(Pascal)/PCIe/SSE2) and one CPU (Intel Core i7), referred to as one resource unit. For each experiment conducted, we use the same resources to train.
Our Othello player model was trained for three days on boards of all sizes, between $5$ and $8$. Our Gomoku player was trained for $2.5$ days on boards of random sizes, between $5$ and $9$. The hyperparameters are selected via preliminary results on small boards. The training parameters for SAZ and the original AZ are presented in Appendix~\ref{supp:ModelParameters}.\footnote{Both the code and the model weights will be available upon acceptance.}

\subsection{Model analysis}
For the model analysis we define some \textit{baseline players}, each trained for three days (unless otherwise specified), as our model was:
\begin{itemize}[leftmargin=*]
    \item \textbf{Model1} refers to training the original AZ (with a CNN replacing the GNN) on the actual board size used for testing.  We used a shallower CNN than the one used in the AZ model, due to our limited computational resources (the architecture is described in Appendix~\ref{supp:cnnArch}). Note that because we failed to train a competitive AZ player with the shallow CNN, we reused symmetries of the training examples (see Section~\ref{trainingPipeline}) as proposed in AGZ model.
    \item \textbf{Model2} refers to training SAZ on the actual board size used for testing, rather than smaller boards.
    \item \textbf{Model3} is the same player as SAZ without the subgraph sampling component, i.e., the action probabilities are taken directly from the output of $f_\theta$ on the full graph.
    \item \textbf{Model4} is the same as SAZ except here we discard the output of $f_\theta$ on the full graph; thus, the action probabilities are calculated only according to the sampled subgraphs' mean.
    \item \textbf{Model5} refers to an MCTS guided by a small CNN. The small CNN was trained by the AZ model on a smaller board of size $m$. The action probabilities are taken as the scatter mean of the network output on all the sub-boards of size $m$ of the state that is evaluated.
\end{itemize}

\paragraph{The merits of our modified components:} 
We start with a small ablation study, where we evaluate the contributions of our main changes. We start with the complete SAZ and leave one component out each time, both for training and realtime playing purposes.
Note that in this experiment, we focus on the first two changed components presented in Section~\ref{description}. Removal of the third component was tested as well, but, as expected, it has no effect on the performance, as the GNN framework has the property of rotation and reflection invariant. It does, however, increase the training time significantly.

Table~\ref{table:ablation} shows the average outcome (see definition in Section~\ref{outcome}) of each model playing against the greedy opponent on a $16\times16$ board for Othello, and $17\times17$ for Gomoku. Blue and red colors represent whether or not a player wins more than $50\%$ of the games against the greedy opponent. In general, it can be seen that removing each component results in a decrease in performance. Both \textbf{model1} and \textbf{model2} produce the poorest results, probably due to insufficient training time on the large board. \textbf{Model3} is already achieving fair results, while our \textbf{SAZ} slightly improves its performance. We will further discuss the subgraph sampling contribution in the next experiment. 

\begin{table}[ht!]
\caption{Leave-one-out study (test average outcome against the greedy opponent)}
\label{table:ablation}
\centering
%\resizebox{0.8\textwidth}{!}{%
\begin{tabular}{ l r r  }
 \\
 \toprule
 \textbf{Model} & \textbf{Othello} $\mathbf{16\times16}$ & \textbf{Gomoku} $\mathbf{17\times 17}$\\
 \midrule
 SAZ [complete model] & \color{blue} $\mathbf{0.85\pm 0.02}$ & $ \color{blue} \mathbf{0.8\pm0.03} $\\
 AZ trained on tested board [model1] & \color{red} $0.44\pm0.02$ & \color{red} $0.43\pm0.04$ \\
 SAZ trained on tested board [model2] & \color{red} $0.32\pm0.03$ & \color{red} $0.34\pm0.04$\\
 only full graph [model3] & \color{blue} $0.78\pm0.02$ & \color{blue} $0.73\pm0.02$ \\
 only subgraphs [model4] & \color{blue} $0.65\pm0.01$ & \color{red} $0.31\pm0.05$ \\
 \bottomrule
\end{tabular}%
\end{table}

\begin{figure}[ht!]
    \centering
    \includegraphics[width=0.98\linewidth]{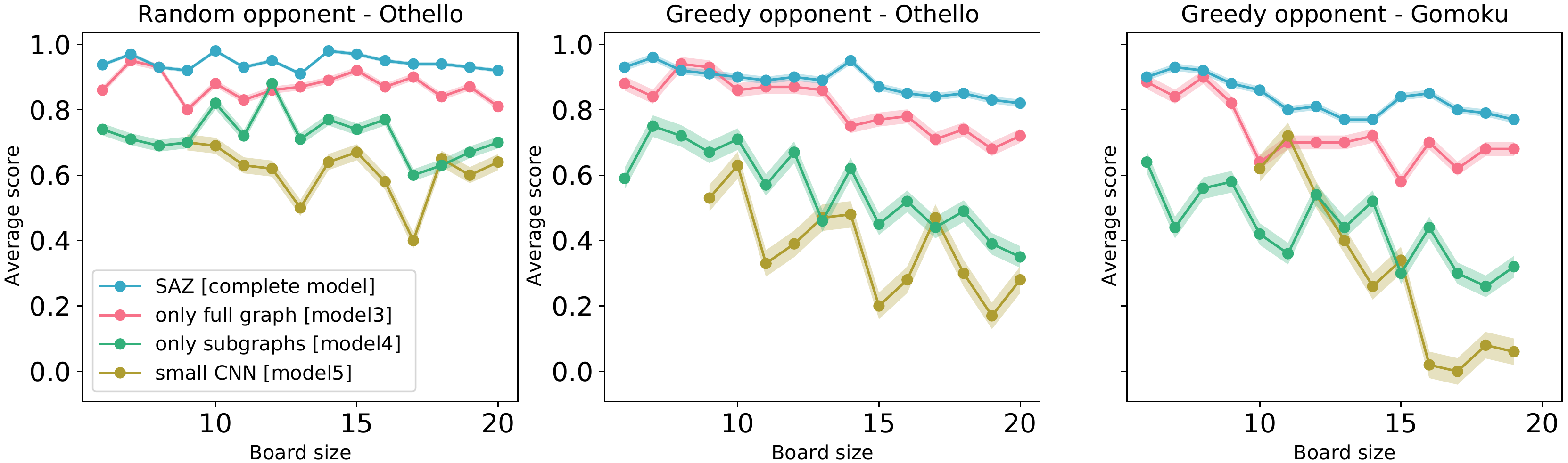}
    \caption{Average outcome of scalable players against the reference opponents on various board sizes and games. The shadowed areas represent the standard errors (5 independent runs).}
    \label{fig:OthelloGeneralization}
\end{figure}

\paragraph{Generalization to larger boards:}
As mentioned, SAZ was designed to allow training and playing on different sizes of input. The generalization study is presented in Figure~\ref{fig:OthelloGeneralization} and shows the average outcome against the reference opponents for Othello and Gomoku, on various board sizes. We also include other baseline players' performance. All models tested in this experiment were trained for three days on our machine. Overall, SAZ performs significantly better than other methods, consistently winning over $75\%$ of the games against the greedy opponent in all cases.

Among all baseline players, \textbf{model4} and \textbf{model5} exhibit the worst performance against both opponents and suffer the greatest performance decrease as the board gets larger. The results of both models suggest that using a small network, applied only on local areas of the full board, does not provide good generalization power, probably because long-term relations are necessary to fully observe the state. \textbf{Model3} is pretty stable along board sizes, reasonably achieving its best results playing on the board sizes on which it was trained. Observe that our Othello \textbf{SAZ} reaches its peak efficacy on a board size that it had not seen during training.

\begin{figure}[ht!]
    \centering
    \begin{subfigure}[b]{0.5\textwidth}
        \centering
        \includegraphics[width = \textwidth]{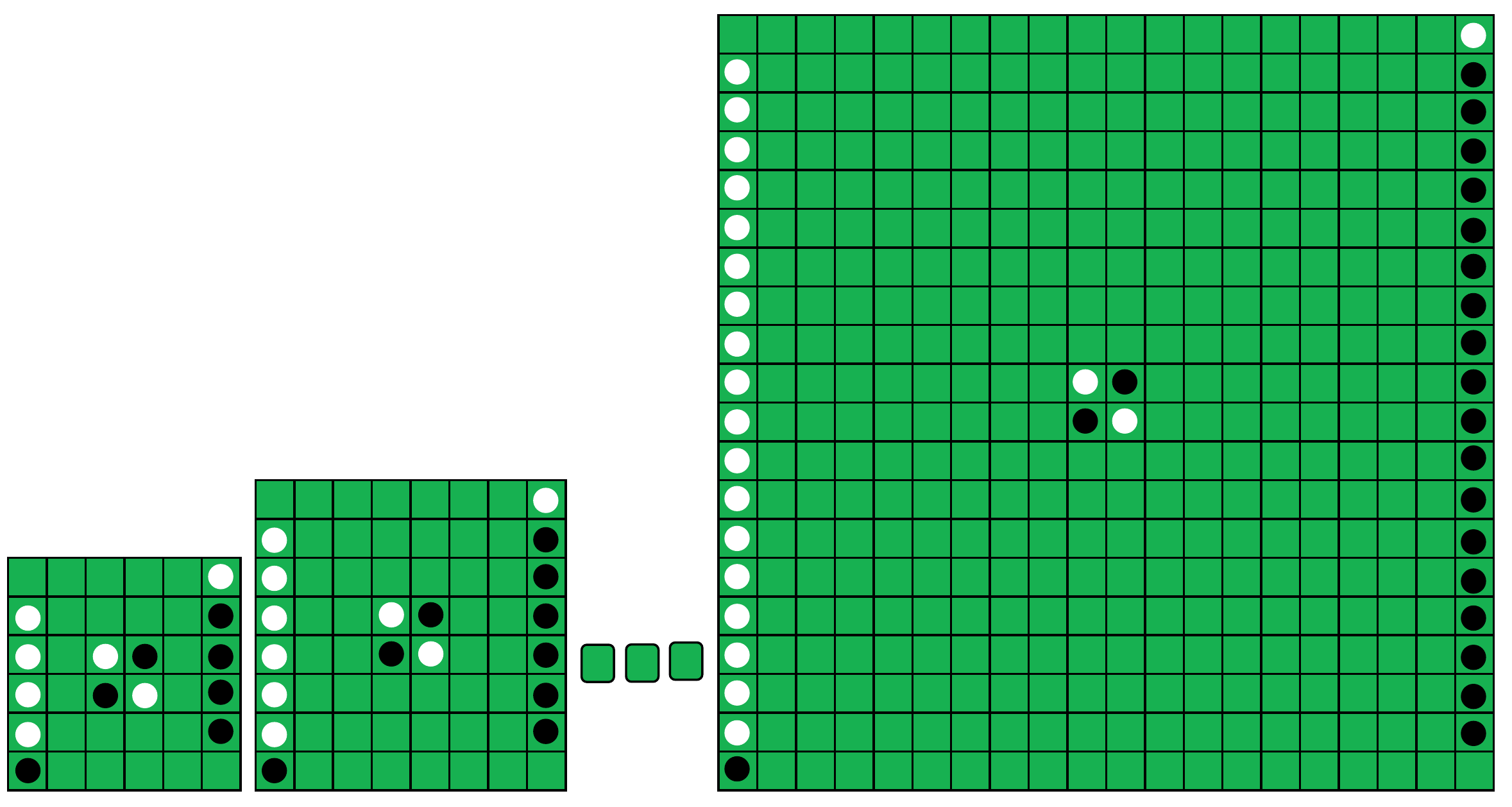}
        \caption{Synthetic Othello boards}
        \label{fig:othelloSynthetic}
    \end{subfigure}
    \begin{subfigure}[b]{0.45\textwidth}
        \centering
        \includegraphics[width = 0.85\textwidth]{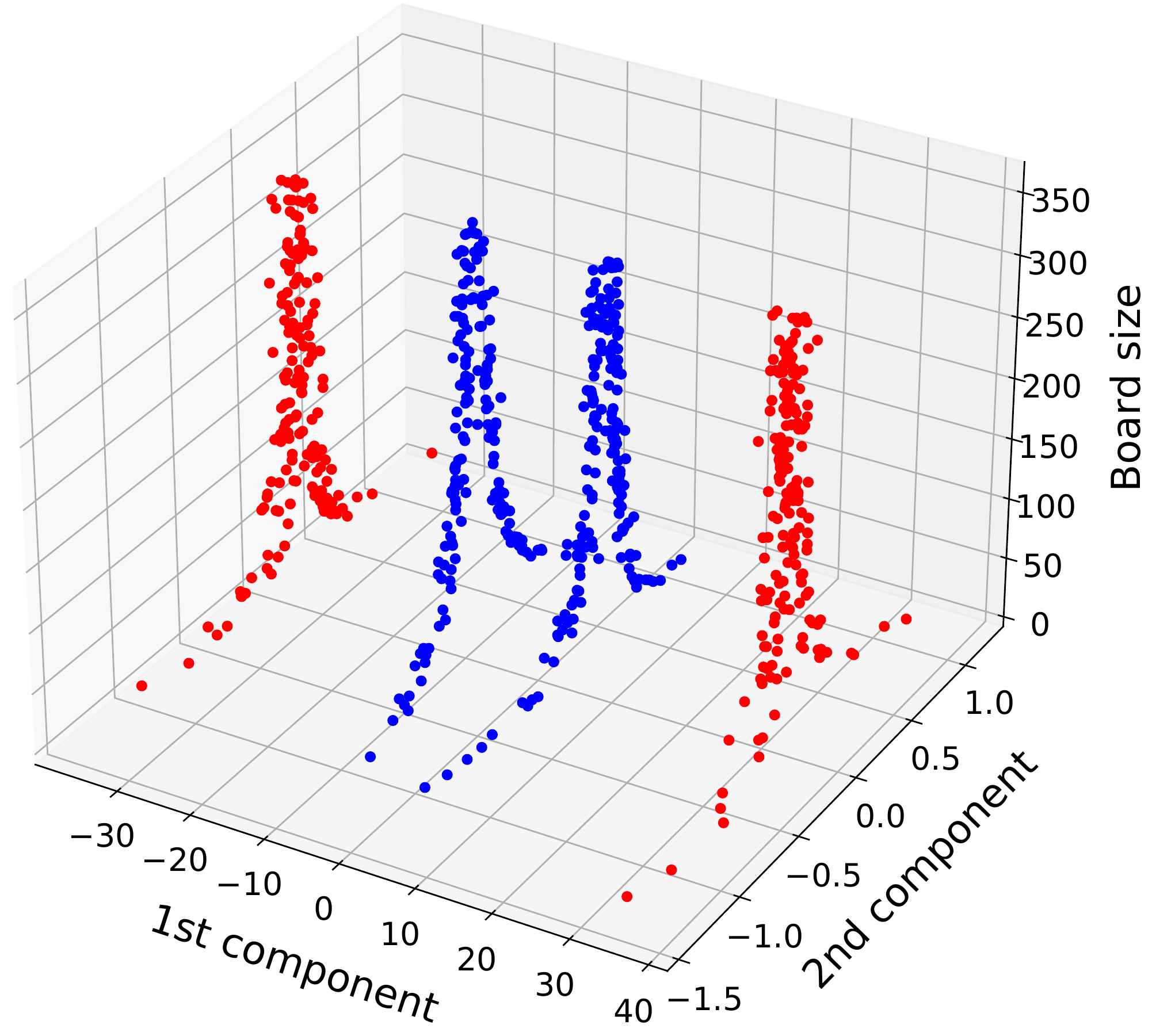}
        \caption{2d PCA projection of good (blue) and bad (red) action embeddings as a function of the board size.}
        \label{fig:action emb}
    \end{subfigure}
    \caption{(a) The synthetic Othello boards of increasing sizes we created. A similar board of the same form was created for all board sizes between $\mathbf{6\times 6}$ and $\mathbf{350\times 350}$. (b) The first two principal components of the embeddings provided by our GNN (X,Y plane) as a function of the board size (Z axis). Blue points refer to the embedding of the ``good action'' of placing a dark piece in the top-left corner. Red ones refer to the ``bad action'' of placing a dark piece in the bottom-right corner.}
    \label{fig:embedding}
\end{figure}

We further examine the generalization power geometrically by considering the GNN actions' latent space. We constructed synthetic Othello boards of specific form, shown in Figure~\ref{fig:othelloSynthetic}, in different sizes from $\mathbf{6\times 6}$ to $\mathbf{350\times 350}$. We apply Principal Component Analysis (PCA) \citep{wold1987principal} on the embedding provided by the GNN for two specific actions -- one that we consider a ``good action'' (top-left corner, capturing all opponent pieces in the first column) and a second that we deem a ``bad action'' (bottom-right corner, which does not capture pieces at all). Figure~\ref{fig:action emb} shows the first two components of the PCA analysis of both actions (on the X,Y plane) as a function of the board size (Z axis). Clearly, except for a few outliers, most of the good actions (blue) are separated easily from the bad ones (red), showing that the latent space successfully encodes the underlying structure of the actions on the board, even for massive board sizes.

\begin{figure}[ht!]
    \centering
    \includegraphics[width=0.85\linewidth]{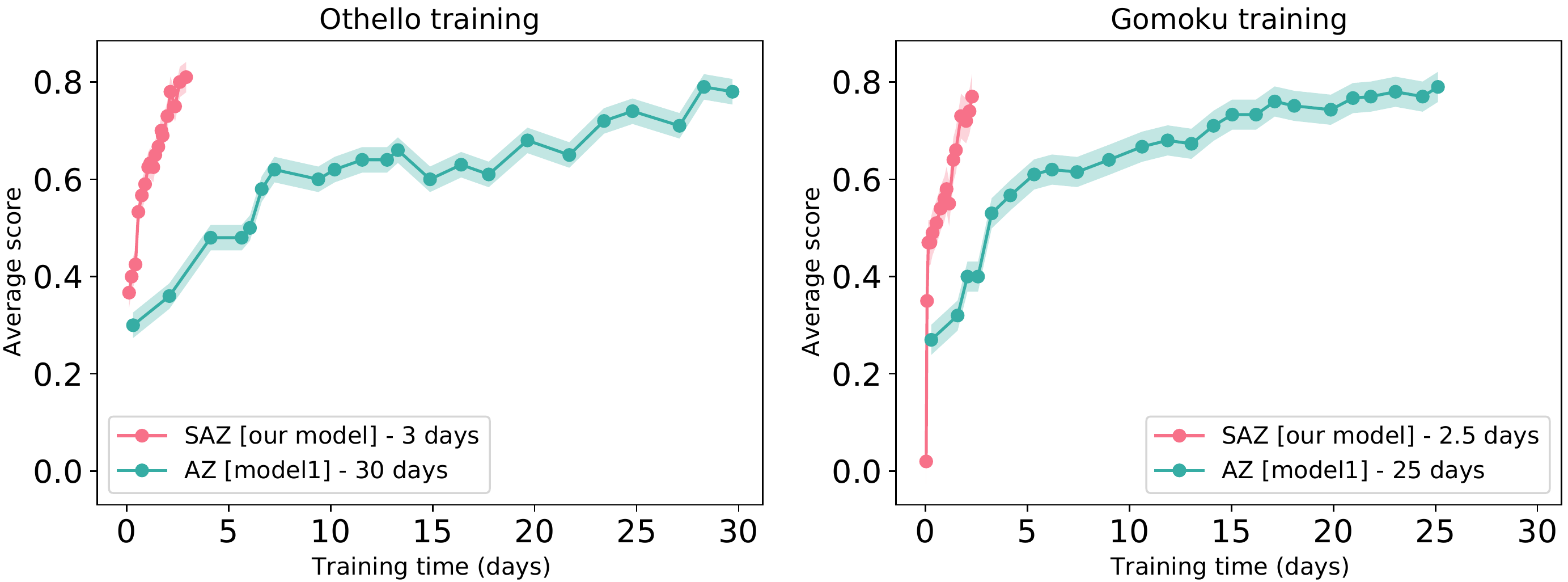}
    \caption{Progression of GNN skill along training. The average outcome is evaluated by playing against the greedy opponent on Othello (board size $16\times 16$) and Gomoku (board size $17\times17$).}
    \label{fig:training}
\end{figure}

\paragraph{Training time analysis:}
Figure~\ref{fig:training} shows the progression of our GNN during training. We measure the GNN \emph{skill} by evaluating the average outcome of \textbf{model3} (i.e., an MCTS guided by the GNN), at each training stage, against the greedy opponent on a $16\times 16$ Othello board and a $17\times17$ Gomoku board. Since we test the GNN on a larger board than the ones used for training, it can be seen as another measure of the generalization power. As a comparison we train \textbf{model1} (i.e., original CNN) on the larger boards for $30$ days and evaluate it along the training time as well.

We observe that as training advances, \textbf{model3} gets stronger, achieving around an $80\%$ win rate at the end of training, and reaching parity with the greedy player after a few hours of training. In contrast, to achieve parity, \textbf{model1} needed between four to five days of training, and achieving \textbf{model3}'s final win rate against the greedy player only after $28$ days (Othello) and $23$ days (Gomoku).

%The same conclusions also hold for the results on Go-Moku (results not shown).

\paragraph{Comparison to AZ:}
Table~\ref{table:AZresults} shows the average outcome of various scalable players (rows) against the original AZ guided by a CNN (columns). Entries in the table represent the average outcome of the game with respect to the row player. Blue and red colors represent whether or not a specific (row) player wins more than $50\%$ of the games against AZ. The scalable players include our model as well as other baseline players, all trained for three days on small boards (up to $9\times 9$). AZ players were trained for $\times 10$ days on the large board of the size that was used for testing ($16\times 16$ or $17\times 17$).

The results show that \textbf{SAZ} wins all competitions, with a more than $50\%$ win rate on Othello and $100\%$ on Gomoku. \textbf{Model3}, which does not use the subgraph sampling technique, also competes fairly well with AZ, but still reduces the performance by $24\%$ on Othello. Both \textbf{model4} and \textbf{model5} Othello players are not competitive compared to AZ, showing again that global dependencies on the board are critical for gameplay. Nevertheless, both models produce a positive win rate against AZ on Gomoku, showing that local structures are more helpful for mastering this game.
To further illustrate the capabilities of SAZ compared to AZ, we conduct the same experiment with $20\times20$ Othello and $19\times 19$ Gomoku boards. The effect is much stronger, as \textbf{SAZ} wins $84\%$ of Othello games against AZ. The AZ $19\times19$ Gomoku player performs poorly in all cases, suggesting that enlarging the board should be accompanied either with a more powerful CNN architecture or with more training.

\begin{table}[ht!]
\caption{Average outcome of scalable players (rows), trained on small boards, against the original AZ players (columns), trained on the tested board size over nearly $\times 10$ more training time.} 
\label{table:AZresults}
\centering
\resizebox{0.8\textwidth}{!}{%
\begin{tabular}{ lrrrr  }
  \\
  \toprule
  & \multicolumn{2}{c}{\textbf{Othello AZ}} & \multicolumn{2}{c}{\textbf{Gomoku AZ}}  \\
  \cmidrule(r){2-3} \cmidrule(r){4-5}
  & \multicolumn{1}{c}{$\mathbf{16\times16}$} &  \multicolumn{1}{c}{$\mathbf{20\times20}$} &  \multicolumn{1}{c}{$\mathbf{17\times17}$} &  \multicolumn{1}{c}{$\mathbf{19\times19}$} \\
  \midrule
  SAZ [complete model] & \color{blue} $\mathbf{0.54\pm0.02}$  & \color{blue} $\mathbf{0.84\pm0.01}$ & \color{blue} $\mathbf{1\pm0.00}$ \color{blue} & \color{blue} $\mathbf{1\pm0.00}$ \\
  only full graph [model3] & \color{red} $0.41\pm0.01$ & \color{blue} $0.72\pm0.01$ & \color{blue} $\mathbf{1\pm0.00}$ & \color{blue} $\mathbf{1\pm0.00}$ \\
 only subgraphs [model4] & \color{red} $0.05\pm0.03$ & \color{red} $0.28\pm0.04$ & \color{blue} $0.55\pm0.4$ & \color{blue} $\mathbf{0.98\pm0.01}$ \\
 small CNN [model5] & \color{red} $0.1\pm0.03$ & \color{red} $0.33\pm0.03$ & \color{blue} $0.74\pm0.05$ & \color{blue} $0.95\pm0.02$ \\
  \bottomrule
\end{tabular}%
}
\end{table}

\paragraph{Go evaluation:} Training AZ to the game of Go with full boards is computationally challenging with our available resources. Recall that Deepmind used $\sim5000$ TPUs for $13$ days to train AZ $19\times19$ Go player. 
We therefore trained our SAZ for three days on Go boards of maximal size $9\times 9$. To test our model we trained two AZ players on boards of sizes $9\times9$ and $15\times15$ for $20$ and $10$ days, respectively. %\footnote{AZ training on $15\times15$ Go board as well as SAZ further parameter tuning are still ongoing.}.
Our analysis suggests that SAZ wins around $68\%$ (on a $9\times9$ board) and $77.5\%$ (on a $15\times15$ board) of the games against AZ. These results as well as the extensive experiments on Othello and Gomoku, which have some similarity to the properties of Go, indicate that our method can lead to solutions that master the game of Go with much less computational overhead.

%While we were also interested in tackling the game of Go, this challenge turned out to beyond our computational resources. Specifically, with the available resources, we were not able to train an RL framework for Go, even on small boards (e.g., $7\times 7$) and using only shallow networks. %Recall that deepmind used xx to train AZ for Go, while we used which did not converge... Othello and Gomoku have some similarity to Go and can be addressed and researched without the need for Google-scale resources. 

\section{Conclusion and future work}
\label{conclusion}
In this paper we presented an end-to-end RL model for training on and playing scalable board games. Central to our approach is the combination of a scalable neural network (GNN), and the AZ algorithm. The use of GNNs facilitated the enhancement of the model by the subgraph sampling technique, and enabled scaling from small boards to large ones. Through extensive experimental evaluation, we demonstrated the effectiveness of our method in learning game strategies, which we validated using different games and various board sizes. The generalization analysis suggests that learning on small boards is faster and more practical than learning solely on large boards. The experiments shown in this paper suggest that SAZ offers a promising new technique for learning to play on large boards, requiring an order of magnitude less training, while keeping the performance level intact. 

We have left a number of potential improvements to future work. First, to date we have focused on board games whose actions refer to the nodes on the graph. This focus was natural because GNNs output the feature vector for each node. Nevertheless, we can use the same approach for another family of board games by using GNNs that estimate edge features (e.g., the game of Chess can be formulated as a graph problem where edges correspond to the actions on the board). A promising approach to achieve this could be to use the method of \cite{berg2017graph} who employ the incident node features to derive edge representations. Furthermore, our subgraph sampling technique, which effectively improved our model performance in our context by reducing the GNN's uncertainty, is of potential independent interest. It would be interesting to validate this approach in different domains. Another promising idea would be to use a model pretrained with our approach and then finetune it to a larger board. The finetuned model would possibly enhance the performance on that size. Finally, it would be important to consider deeper GNN architectures, which will possibly enable discovering longer term dependencies on the board.

%Finally, as we were not able to train the model on the game of Go, due to lack of computing resources, it would be important in the future to validate our approach on Go using sufficient computation power. 

% TODO: use order of appearance or alphabetically?

\bibliographystyle{unsrtnat}
\medskip
\bibliography{references}

\newpage
\appendix

\section{AlphaZero}
\label{supp:AZ}
As we mentioned in the Section~\ref{related:RL}, \cite{silver2017mastering_2} proposed an RL algorithm for board game playing. It uses a neural network $f_\theta$, which is used for guiding the internal steps of an MCTS. $f_\theta$ gets as input a state $s$ and outputs a probability vector for all possible moves $\mathbf{p}_s$, and a scalar $v_s\in [-1,1]$, which corresponds to the network's confidence regarding the current player's chances winning the game.

\subsection{Monte Carlo tree search:} 
\label{supp:MCTS}
The tree search is designed to explore the game states and actions, and to provide an improved probability vector $\pi_s$. Here we describe the MCTS variant used in the AZ framework.
For each pair $(s,a)$, corresponding to the state and action, it stores the following variables:
\begin{itemize}
    \item $Q(s,a)$: The action value.
    \item $P(s,a)$: The probability of choosing $a$ from the state $s$
    \item $N(s,a)$: The visit count of the pair $(s,a)$.
    \item $U(s,a)$: Upper confidence bound of the pair $(s,a)$ computed by: 
            $$U(s,a)=c_{\textrm{puct}}P(s,a)\frac{\sqrt{\sum_{a'\in \mathcal{A}}N(s,a')}}{1+N(s,a)}$$ where $c_{\textrm{puct}}$ is a hyperparameter that controls the exploration and exploitation.
\end{itemize}

Each round of MCTS consists of:
\begin{enumerate}
    \item \textbf{Selection:} Start at the root $s_0$ and select a child node maximizing $Q(s,a)+U(s,a)$ until an unexpanded node $s'$ is reached.
    \item \textbf{Expansion}: If $s'$ is a terminal state (i.e., has a decisive result $z\in \{-1,0,1\}$, a win, a tie or a loss), let $v_{s'}=z$. Otherwise, evaluate $f_\theta(s')=(\mathbf{p}_{s'},v_{s'})$ and store $P(s',\cdot)=\mathbf{p}_{s'}$
    \item \textbf{Backpropagation}: Traverse all the pairs $(s,a)$ visited along the path to $s'$ and update:
    \begin{equation}\label{eq:Qsa}
    Q(s,a) = \frac{N(s,a)\cdot Q(s,a)+v_{s'}}{N(s,a)+1}
    \end{equation}
    \begin{equation}\label{eq:Nsa}
    N(s,a) = N(s,a)+1
    \end{equation}
\end{enumerate}

After a predefined number of rounds, calculate the improved probability vector $\pi_{s_o}$. The vector element in the location corresponding to the action $a$ is:
$$\pi_{s_0}^a=N(s,a)^{\nicefrac{1}{\tau}}$$
where $\tau$ is a temperature parameter. When $\tau$ is large, the probability vector is much closer to a uniform distribution; when $\tau\rightarrow 0$, the probability of the most visited action is closer to $1$. Usually, we reduce $\tau$ as the learning advances.

\subsection{Training pipeline:} 
\label{supp:pipeline}
The training is composed of a loop between two independent stages:
\begin{itemize}
    \item \textbf{Selfplay:} The player plays against itself, using MCTS guided by the latest weights of $f_\theta$. The selfplay accumulates training examples of the form $(s,\pi_s,z_s)$, where $s\in\mathcal{S}$ is the state (usually in a canonical form), $\pi_s$ is the probability vector obtained from MCTS and $z_s$ is the final result  of the game (when using the canonical form for $s$, we always take the perspective of a specific player).
    At the end of this stage, AZ updates the training set to include all the boards that can be constructed by a rotation or reflection of an example in the training set.
    \item \textbf{Optimization:} After constructing the training set in the previous stage, the neural network is trained to maximize the similarities between $\mathbf{p}_s$ and $\pi_s$, and to minimize the difference between $v_s$ and $z_s$. The loss function used to achieve this goal (for a single example) is:
    $$ \mathcal{L}(s) = (z_s-v_s)^2+\textrm{CrossEntropy}(\mathbf{p}_s,\pi _s)+c||\theta||^2_2$$ where $c$ is a regularization factor.
\end{itemize}
The training examples are kept between iterations. When one iteration ends, the oldest training examples are partially removed.

\section{Graph neural networks}
\subsection{Message passing procedure}
\label{supp:MP}
The \textit{message passing} algorithm is a central component of graph neural networks.
It uses a predefined number of iterations to propagate information between nodes on the graph. Here we describe it in details.
In its basic form, the message passing algorithm receives as input a graph $G=(V,E)$ and the number of overall iterations $T$, and stores hidden representations $h_k^t\in d_t$ of the graph nodes, where $k\in V$, $t\in \{1,\cdots,T\}$ and $d_t$ is the hidden dimension of layer $t$.

At iteration $t$, each node $k$ receives messages from its graph neighbors, denoted by $N(k)$. Messages are generated by applying a message function $m(\cdot)$ to the hidden states $h_i^t$ of nodes in the graph, and then are combined by an aggregation function $\textrm{AGG}(\cdot)$, e.g., a sum or a mean (Equation~\ref{eq:GNN_agg}). An update function $u(\cdot)$ is later used to compute a new hidden state $h_k^{t+1}$ for every node $k$ (Equation~\ref{eq:GNN_updaterule}). Finally, after $T$ iterations, a readout function $g(\cdot)$ outputs the final prediction, based on the final node embeddings $h_k^T$ (see Equation~\ref{eq:GNN_pred_node} for node prediction and Equation~\ref{eq:GNN_pred_graph} for graph prediction). Neural networks are often used for both $m(\cdot)$, $u(\cdot)$ and $g(\cdot)$.

\begin{equation}\label{eq:GNN_agg}
M_k^{t+1}=\underset{i\in N(k)}{\textrm{AGG}}\ m(h_k^t,h_i^t)
\end{equation}
\begin{equation}\label{eq:GNN_updaterule}
h_k^{t+1}=u(h_k^t,M_k^{t+1})
\end{equation}
\begin{equation}\label{eq:GNN_pred_node}
o_k=g(h_k^T,h_k^0)
\end{equation}
\begin{equation}\label{eq:GNN_pred_graph}
o_G=g(\underset{k\in V}{\textrm{AGG}}\ h_k^T)
\end{equation}

\subsection{Graph isomorphism networks}
\label{supp:GIN}
\citeauthor{xu2018powerful} proved that the graph isomorphism network (GIN) model is as powerful as the Weisfeiler-Lehman graph isomorphism test and is the most expressive among the class of GNNs. We describe a hidden feature update layer of GIN, from a message passing perspective. At iteration number $t$, each node $k$ is updated by:
$$h_k^{t+1}=h_\theta\left(\left(1+\epsilon\right)h_k^t+\sum_{j\in N(k)}h_j^t\right)$$
where node features are aggregated by a summation operation, $\epsilon$ is either a learnable parameter or a fixed scalar and $h_\theta$ denotes a neural network (i.e., an MLP). The same update rule can be computed in a matrix form as:
$$H^{t+1}=h_\theta \left( \left( A + \left( 1 + \epsilon \right) \cdot I \right) \cdot H^t \right)$$
where $A$ is the adjacency matrix of $G$ and $I$ is the identity matrix.
Note that in our GNN architecture we used a two headed network for computing the policy (node regression task) and the value (graph classification task). 

%The general idea behind graph convolutional neural networks (GCNs) \citep{defferrard2016convolutional} is to apply convolution over a graph. We describe a hidden feature update layer of GCN, from a message passing perspective. At iteration number $t$, each node $k$ is updated by:
%$$h_k^{t+1}=\sum_{i\in N(k)\bigcup \{k\}}\frac{1}{\sqrt{\deg(k)\deg(i)}}(\Theta^{t+1}\cdot h_i^{t})$$
%where neighboring node features are first transformed by a weight matrix $\Theta$, normalized by their degree, and finally aggregated by a summation operation. The same update rule can be computed in a matrix form as:
%$$H^{t+1}=\hat{D}^{-\nicefrac{1}{2}}\hat{A}\hat{D}^{-\nicefrac{1}{2}}H^t\Theta^{t+1} $$ where $\hat{A}=A+I$ is the adjacency matrix of $G$, with inserted self loops, and $\hat{D}$ is the corresponding diagonal degree matrix.

\section{Pseudocode}
\label{supp:Pseudocode}
Here we provide the pseudocode for our ScalableAlphaZero model. Algorithm~\ref{alg:MCTSsearch} describes the MCTS parameters update starting from an initial state $s$, Algorithm~\ref{alg:getPi} describes the update rule for the policy vector $\pi$ based on the updated MCTS and Algorithm~\ref{alg:train} describes the training pipeline.

\begin{algorithm}[htp!] 
\SetAlgoLined
\KwIn{an initialized MCTS tree $\mathcal{T}$, a state $s_0$ (root), number of subgraph to use $k$, size of subgraphs $m$.}
$s\leftarrow s_0$ \\
\uIf {$s$ is terminal} {
    $v_s\leftarrow$ get game result of $(s)$ \\
}
$g\leftarrow $ convert $s$ to graph \\
\While{$s$ is not terminal}{
    \uIf{$s$ is not expanded} {
        $(\mathbf{p}_1,v_s) \leftarrow GNN(g)$ \\
        sample $k$ subgraphs of size between $(m-1)^2$ and $(m+1)^2$ \\
        $(\mathbf{p}_2,\sim) \leftarrow GNN(\textrm{subgraphs})$ \\
        $Q(s,a)\leftarrow v_s$ \\
        $P(s,\cdot) \leftarrow \frac{p_1+p_2}{2}$ \\
        $N(s,\cdot)\leftarrow 0$
    }
    $a\leftarrow$ action that maximizes $U(s,\cdot)$\\
    $s\leftarrow$ next state of $s$ after selecting $a$
}
\While{$s\neq s_0$}{
    $a\leftarrow$ previous action \\
    $s\leftarrow$ previous state \\ 
    $Q(s,a)\leftarrow \nicefrac{\left(Q(s,a)\cdot N(s,a)+v_s\right)}{\left(N(s,a)+1\right)}$ \\
    $N(s,a)\leftarrow N(s,a)+1$
}
\KwOut{an updated $\mathcal{T}_s$.}
\caption{$\textrm{MCTS}_\tau$}
\label{alg:MCTSsearch}
\end{algorithm}

\begin{algorithm}[htp!] 
\SetAlgoLined
\KwIn{a state $s$, a temperature $\tau$, the number of MCTS simulations $N_\textrm{sim}$.}
\For{$i=1$ to $N_\textrm{sim}$}{
    $\textrm{MCTS}_\tau(s)$
}
$\pi(s,\cdot)\leftarrow N(s,\cdot)^{\frac{1}{\tau}}$ \\
\KwOut{$\pi_s$}
\caption{compute $\pi_s$}
\label{alg:getPi}
\end{algorithm}

\begin{algorithm}[htp!] 
\SetAlgoLined
\KwIn{maximal board size for training $n$ (squared), the number of AZ iterations $N_{\textrm{iter}}$, a GNN $f_\theta$, an MCTS $\mathcal{T}$, the number of AZ iterations to include in history $H$.}
\For{$i=1$ to $N_\textrm{iter}$}{
    sample board size $n_0\in\{n,n-1,n-2,n-3\}$ w.r.t the probability vector $(0.4,0.3,0.2,0.1)$ \\
    \emph{training examples} $\leftarrow$ selfplay($n_0\times n_0,f_\theta,\mathcal{T}$) \\
    add \emph{training examples} to \emph{history} \\
    \uIf{length(\emph{history})>$H$}{
        pop history
    }
    \For{batch in shuffled \emph{history}}{
        $(\mathbf{Ps},Vs) \leftarrow GNN(\textrm{batch})$ \\
        $\textrm{batchloss} \rightarrow MSE(z(\textrm{batch}),Vs)+CrossEntropy(\pi_\textrm{batch}, Ps$) \\
    }
}
compute total loss \\
optimize GNN parameters $\theta$ to minimize total loss \\
\KwOut{optimized GNN $f_\theta$}
\caption{train}
\label{alg:train}
\end{algorithm}

\section{Global setup}
\label{supp:ModelParameters}

\paragraph{GNN:} We used three layers of GIN with a $ReLU$ nonlinearity and a hidden dimension of $512$.

\paragraph{MCTS:} The number of MCTS simulations was set to $N_\textrm{sim}=100$. We used $c_\textrm{PUCT}=1.5$ for the exploration and exploitation parameter. The temperature was set to $\tau=1$ at the beginning of the tree search and, after $25$ search iterations, was changed to $\tau=0$ (i.e., the action is chosen by \textit{argmax}).
Consider a board of size $n\times n$. The number of sampled subgraphs is larger when the board size increased and is set to $\textrm{round}(n/2)$. For the parameter that controls the subgraphs size we used $m=n-1$ or $m=n-2$.

\paragraph{CNN architecture:}
\label{supp:cnnArch}
CNN architecture is relevant to the experiments that include the original AZ player (i.e., \textbf{model1}).
It contains the following modules:
\begin{enumerate}[leftmargin=*]
    \item 2d convolutional layers with $512$ channels, a kernel of size three, stride$=1$ and padding$=1$, followed by 2d batch normalization and a $ReLU$ activation function.
    \item 2d convolutional layers with $512$ channels, a kernel of size three and stride$=1$, followed by 2d batch normalization and a $ReLU$ activation function.
    \item A fully-connected layer with hidden dimension of size 1024 and dropout, followed by 1d batch normalization and a $ReLU$ activation function.
    \item A fully-connected layer with hidden dimension of size 512 and dropout, followed by 1d batch normalization and a $ReLU$ activation function.
    \item The computation is separated into two different heads, for computing the policy $\mathbf{p}$ and the value $v$. $\mathbf{p}$ is computed using one fully-connected layer from input of size $512$ to output of size $\left|\mathcal{A}\right|$ (number of possible actions), followed by a $\log$-$softmax$ operation, yielding the probability vector. $v$ is computed using one fully-connected layer from input of size $512$ to output of size $1$, followed by a $\tanh$ nonlinearity function.
\end{enumerate}

\paragraph{Greedy players heuristics:}
\label{supp:heuristics}
As mentioned in Section~\ref{Evaluation}, for our challenging baseline opponent we defined a greedy player, which chooses his actions based on a hand-coded heuristic score. The heuristics are unique for each game: for the game of Othello, the state score is the difference between the player's stones and those of his opponent; for the game of Gomoku, the score is the length of the maximal sequence of the current players' stones minus the length of the maximal sequence of opponents' stones; for the game of Go, the score is evaluated by the difference between the player's territories and those of his opponent.

\paragraph{Training and environment:}
\label{supp:env}
Our loss function did not include a regularization term (i.e., $c=0$).
The training set included examples from $20$ iterations of selfplay and optimization. 

For our multiple-sized SAZ training we randomly sampled a board size at the beginning of each game in the selfplay procedure (see Section~\ref{supp:AZ}), taken from a probability distribution that is proportional to the board size. For example for training GoMoku we used boards of sizes $(6\times6,7\times7,8\times8,9\times9)$ and the probability vector for choosing each size was $(0.4,0.3,0.2,0.1)$. The full algorithm is described in Section~\ref{supp:Pseudocode}.

We used PyTorch Geometric \citep{fey2019fast} for the implementation of the GNN. We used alpha-zero-general for the re-implementation of the AlphaZero model with our modified components,\footnote{GitHub repository: \href{https://github.com/suragnair/alpha-zero-general}{alpha-zero-general} (released under the MIT license).} and used the Go environment from alpha-zero-general-with-go-game.\footnote{GitHub repository: \href{https://github.com/joelmichelson/alpha-zero-general-with-Go-game}{alpha-zero-general-with-go-game} (released under the MIT license).}

%%%%%%%%%%%%%%%%%%%%%%%%%%%%%%%%%%%%%%%%%%%%%%%%%%%%%%%%%%%%

\end{document}